\documentclass[conference]{IEEEtran}
\IEEEoverridecommandlockouts
\usepackage{cite}
\usepackage{amsmath,amssymb,amsfonts}
\usepackage{algorithmic}
\usepackage{graphicx}
\usepackage{textcomp}
\usepackage{xcolor}
\usepackage{tikz}

\newcommand\copyrighttext{%
  \footnotesize \textcopyright 2023 IEEE. Personal use of this material is permitted. Permission from IEEE must be obtained for all other uses, in any current or future media, including reprinting/republishing this material for advertising or promotional purposes, creating new collective works, for resale or redistribution to servers or lists, or reuse of any copyrighted component of this work in other works. }
\newcommand\copyrightnotice{%
\begin{tikzpicture}[remember picture,overlay]
\node[anchor=south,yshift=10pt] at (current page.south) {\fbox{\parbox{\dimexpr\textwidth-\fboxsep-\fboxrule\relax}{\copyrighttext}}};
\end{tikzpicture}%
}

\def\BibTeX{{\rm B\kern-.05em{\sc i\kern-.025em b}\kern-.08em
    T\kern-.1667em\lower.7ex\hbox{E}\kern-.125emX}}
\begin{document}

\title{Evaluating Adversarial Robustness with Expected Viable Performance
\thanks{Research funded by AFRL}
}



\author{\IEEEauthorblockN{Ryan McCoppin}
\IEEEauthorblockA{\textit{Air Force Research Laboratory} \\
\textit{CAE USA}\\
WPAFB, Ohio, USA \\
ryan.mccoppin.1.ctr@afrl.af.mil}
\and
\IEEEauthorblockN{Colin Dawson}
\IEEEauthorblockA{\textit{Air Force Research Laboratory} \\
\textit{Aptima, Inc.}\\
WPAFB, Ohio, USA \\
cdawson@aptima.com}
\and
\IEEEauthorblockN{Sean M. Kennedy, Leslie M. Blaha}
\IEEEauthorblockA{\textit{711$^{th}$ Human Performance Wing} \\
\textit{Air Force Research Laboratory}\\
WPAFB, Ohio, USA \\
\{sean.kennedy.10, leslie.blaha\}@afrl.af.mil}

}

\maketitle

\copyrightnotice

\begin{abstract}

We introduce a metric for evaluating the robustness of a classifier, with particular attention to adversarial perturbations, in terms of expected functionality with respect to possible adversarial perturbations. 
A classifier is assumed to be non-functional (that is, has a functionality of zero) with respect to a perturbation bound if a conventional measure of performance, such as classification accuracy, is less than a minimally viable threshold when the classifier is tested on examples from that perturbation bound. 
Defining robustness in terms of an expected value is motivated by a domain general approach to robustness quantification.

\end{abstract}


\section{Introduction}
\label{sec:intro}
\begin{figure}[ht]
        \centering
        \includegraphics[width=8cm]{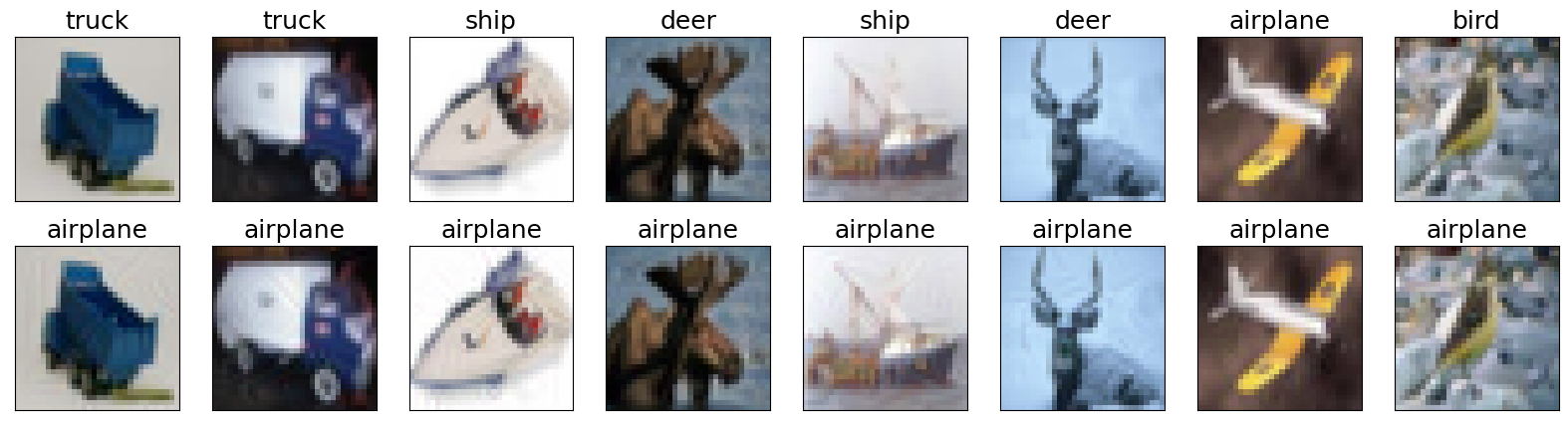}
        \caption{Samples from the CIFAR dataset (top). Images altered using adversarial perturbations (Projected Gradient Descent) which are imperceptible to the human eye, but which alter ResNet50's classification label (bottom).}
      \label{fig:evasion}
    \end{figure}

In recent years, adversarial machine learning (ML) has become a threat to model security and reliable performance. 
Adversarial ML arises when some aspect of the system is intentionally manipulated to cause the classifier to make errors.
Adversarial robustness specifically seeks to measure a model's performance when these perturbations are chosen selectively to be maximally disruptive. 
For example, evasion attacks add human-imperceptible perturbations to a data instance to alter the output of a classifier, as illustrated in Fig.~\ref{fig:evasion}.
Because humans do not make the same classification errors as the model on these images, evasion attacks result in loss of trust in the data and, in turn, loss of trust in the models~\cite{overcoming2023}. 
Adversarial perturbations in the evasion space are only noticed after model performance is harmed. 
The consequence of this is that the classifier will be unreliable and therefore unusable for its intended purpose.
Thus, it has become a significant task to measure the trustworthiness of a model, which necessitates evaluating the robustness to adversarial attacks.

Adversarial robustness highlights that when we want to put ML into use in the world, we expect it to operate successfully in variable environments.
That environment will involve natural sources of noise (e.g., weather will impact lighting in images) and intentional adversarial attacks on the system.
It is desirable for a system to be able to continue to perform against both sources of variance.
To have confidence that ML will operate successfully in the presence of perturbations, we need to define several things about the system: the level of performance that is considered acceptable, the context or environment in which the system will be deployed, the sources of perturbation relevant in that context, and the mechanisms that may provide robustness in performance against those perturbations.
In support of this, it is desirable to have an approach to quantifying robustness that applies generally across various choices to all these system elements.

We adopt the perspective that robustness is the ability of an ML system to maintain its functionality at an acceptable level of performance when some aspect of the system is subject to perturbation, consistent with the domain-general definition laid out by~\cite{gluck}.
From this perspective, robustness is a continuum in a multivariate space, rather than a binary system property.
This allows us to quantify the degree to which robustness changes as different aspects of the system are varied.



Walsh and colleagues~\cite{walshq,walsh} noted that to quantify robustness, the functionality of a system must first be defined in the context of  the system's use;
    functionality may be defined by one or more performance measures for the system.
Classifier performance can be measured in a variety of ways. 
The choice of which metric to use to evaluate a given classification model depends heavily on its use case. 
For some applications, minimizing false positives is of paramount importance;
  for others, errors can be treated more symmetrically, making overall accuracy a suitable metric.
Robustness quantification can be defined as a function of any of these metrics alone or in combination.
  
Robustness of ML systems requires that models exhibit ``graceful degradation'' of performance as inputs to the model are perturbed, whether by natural distortion, distribution shift, or adversarial manipulation. 
Robustness has often been measured in terms of model resilience to acceptable levels of perturbation~\cite{evaluating} or levels of performance in the perturbation space~\cite{benchmarking}. 
More recent metrics take into account model properties such as Lipschitz continuity~\cite{spade} or the interpretability of the generated attention map~\cite{connection}. 
Most robustness metrics involve evaluating model performance on test data consisting of samples perturbed up to a defined magnitude, which may provide only a limited interpretation of the system robustness when operating in variable environments. 

In this paper, we propose a general approach to measuring system robustness as a continuous function of ML performance under any type and degree of perturbation, to include adversarial attacks.
This paper is organized as follows. 
We review the related work on measuring ML robustness.
Next, we introduce the concept expected viable performance (EVP) for considering ML systems in context, and we define a robustness metric in terms of this functionality.
We illustrate the use of EVP robustness on publicly available models.
We end with discussion of how the EVP approach may be used to inform further research on mechanisms to support robust ML systems.


\section{Related Work}
\label{sec:background}


Machine learning robustness assessment approaches fall into four basic categories of metrics: model based, test-data based, perturbation-distance based, and hybrid approaches. 
We look at each category next.

\subsection{Model-Based Robustness}
Model-based (or ``white box'') metrics use properties of a model to determine robustness. 
Test data may be used tangentially to calculate gradients or similar, but it is the gradients, properties of the model or latent space that determine robustness instead of prediction statistics.

The CLEVER Score~\cite{clever} was proposed around the idea of Lipschitz continuity in a model. 
The idea is that models with a ``shallow gradient'' with respect to the input space around the decision boundary would need a larger perturbation to change the class value. 
A sharp gradient between classes means that a small perturbation can cause a large change in the output prediction. 
CLEVER posits that if small movements cannot change the class, the model must be more robust. 
However clever the metric may be,~\cite{goodfellow} showed it to be inadequate in some conditions, using gradient masking to provide examples of successful attacks for which the CLEVER score failed to penalize the model.

Similarly, a recent line of work has instead focused on certificates of robustness. A certificate of robustness probabilistically guarantees that no adversarial examples exist within a ``certified region'' for a specific class of model~\cite{lee2019tight}. Unfortunately, obtaining exact guarantees can be computationally intractable~\cite{lee2019tight}. Another issue with this approach is that some small changes may be semantically meaningful while not all small perturbations are adversarial examples~\cite{tramer2020fundamental}.

Most metrics in this category require access to model properties in their calculations. 
The SPADE score does not, though it approximates a model property.
The SPADE score~\cite{spade} is a spectral method for black box adversarial robustness evaluation that provides an alternative to gradient-based approaches, using Spectral Graph Theory to compute an upper bound for the global Lipschitz constant. 
Because SPADE does not require model gradients, it can be considered a ``black box'' approach, making it more broadly applicable. 

\subsection{Test Data-Based Robustness}

Test data-based (or ``black box'') robustness metrics rely on test data and model output alone, without invoking model properties or parameters. 
The most straightforward is Adversarial Accuracy, which is simply the proportion of adversarially generated test instances with a particular perturbation budget that are classified correctly.
Adversarial Accuracy can be augmented into the Accuracy-Perturbation Curve which visualizes the gradual degradation of a model's performance with stronger attacks~\cite{benchmarking}. 

The accuracy-robustness Pareto frontier~\cite{pareto} describes a set of solutions measured in terms of both accuracy and a known robustness metric. 
This reveals the compromise between robustness and accuracy, a well known problem in the field \cite{oddswith}. 
This tradeoff is a natural consequence of the clean model optimizing for generalization to clean data, such that altering the optimization criterion prevents the model from finding as high a peak. 
The issue is worsened by the potential for adversarial training overfitting to a specific degree of perturbation~\cite{oddswith}. 
One proposed method for combining clean performance and robustness is the Self-COnsistent Robust Error (SCORE)~\cite{proper}.

\subsection{Perturbation distance-based Robustness}


Another set of metrics measure how large an optimized perturbation needs to be to alter the output of the classifier, with larger distances representing a more robust classifier. 
This is generally accomplished by gradually creating adversarial examples using Projected Gradient Descent (PGD)~\cite{pgd} until the sample's class prediction changes. 
The moment the prediction is changed, the perturbation distance is recorded for that example~\cite{evaluating}. 
The average minimum perturbation distance is one of the original definitions of adversarial robustness~\cite{evaluating} and is inversely related to robustness.

\subsection{Hybrid metrics}
The last group of metrics combine properties of test-data based metrics and perturbation-distance based metrics. 
It is in this group that the metric we propose in Section~\ref{sec:ours} falls.

The ARA metric~\cite{woods} measures aggregated classification accuracy over a range of perturbation sizes.
It measures the Area between the Accuracy-Robustness curve and that of a na\"{i}ve (random) classifier whose accuracy is the constant reciprocal of the number of classes.
ARA represents an aggregate measure of above-chance performance across a range of possible perturbation magnitudes. 
\cite{woods} use Euclidean distance normalized by the square root of the number of pixels in the attack (i.e., the $L_2$ norm divided by the square root of the $L_0$ norm) to measure perturbation magnitude.

\begin{figure}[t]
        \centering
        \includegraphics[width=8cm]{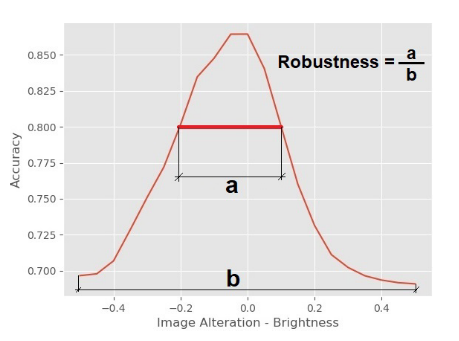}
        \caption{Diagram of the ROBY metric~\cite{robytool}. A threshold value (red horizontal line) in the performance determines a functional region in accuracy versus perturbation space. }
      \label{fig:roby}
    \end{figure}

Similarly, the ROBY analysis tool~\cite{robytool} considers an interval of ``plausible alterations''~\cite{robytool} around an image.
This is depicted in Fig.~\ref{fig:roby} by the interval of length $b$. 
\cite{robytool} also defined a minimum performance threshold, denoted $\theta$, represented by the height of the horizontal red line.
The interval of potential perturbations in which the target performance statistic exceeds $\theta$ (the interval with length $a$) is compared to the larger interval representing all ``plausible alterations'', and the robustness statistic is the ratio of their lengths in perturbation space. 
In Fig.~\ref{fig:roby}, the result is the ratio $a/b$. 
This statistic requires a consistent maximum perturbation size of $b$, which must be constant across all comparisons for ROBY to be informative.

ARA and ROBY metrics are consistent with our perspective that robustness should be measured as a function of performance over a range of perturbations.
The ARA is calculated as an area under the Accuracy-Perturbation Curve, defining accuracy as the performance metric of interest. 
But it relies on the area under the entire curve above a na\"{i}ve baseline, rather than being informed by the planned applications.
ROBY uses the concept of viable performance, but it requires specification of
a maximum perturbation size.
This can introduce user degrees of freedom and variability across experiments. 
In the next section, we motivate and define a new metric designed to overcome these limitations.


\section{Expected Viable Performance}
\label{sec:ours}
For a system to be useful, it must meet a minimum threshold of functionality. 
Increases in performance above that threshold represent gains (in efficiency, effectiveness, etc.). 
On the other hand, distinctions among levels of performance below the threshold in a given context are practically irrelevant if the system will not be used due to not meeting the functional requirements. 
In biological and engineering systems, the imperative of maintaining performance above a threshold is evident as consistent sub-threshold performance could impact individual or species survival or lead to collapse of structures.    

In many classification applications, the classifier must achieve performance rates above a minimum threshold to be considered viable for that application.
For example, a handwriting classifier, like the method by~\cite{kim1997lexicon} used in postal mail sorting, could save a business overhead in terms of person-hours spent on that task, but an error rate greater than 5\% will lead to costly mistakes that overtake the person-hour savings.  
In this example, a viability threshold can be defined as a error rate of less than or equal to 5\%.
When assessing the relative robustness of two classifiers for a range of perturbation regions, it may make sense to ignore distinctions in performance below the threshold of viability.

In the work of~\cite{intriguing}, robustness is defined as the minimal perturbation required to flip the predicted labels. 
While this is a reasonable measure of robustness with respect to a specific natural input being manipulated intentionally via an adversarial attack, it is not necessarily well suited to measuring robustness in the context of {\em natural} sources of perturbations.
This definition fails to distinguish between a system that maintains a high (but not perfect) level of accuracy above a perturbation threshold and a system that degrades more severely;
 the first system is intuitively more robust, and may achieve acceptably low error rates, while the second does not. 
We argue that both performance inside the region of viability and the size of that region ought to be taken into account when measuring robustness. 

\cite{walshq} quantify robustness as the expected value of a system's functionality with respect to possible perturbations;
  importantly, the notion of functionality is defined separately from the perturbations and the metrics of robustness. 
Formally, robustness $R$ of system $s$ is the integral of a performance metric with respect to a probability distribution over potential perturbations: 
\begin{equation}
\label{walshequation}
R(s) = \mathbb{E}[f(s,\delta)] = \int f(s,\delta)\ p(\delta) \,d\delta
\end{equation}
where $f(s,\delta)$ is a measure of the functionality of the system in an environment with perturbations of magnitude $\delta$.

We propose a new ML robustness metric which takes this form, which we term Expected Viable Performance (EVP).

\subsection{Motivation and Definitions}

Because a system is, by definition, not useful in a environment where performance falls below a threshold of viability, we define functionality $f$ of a system $s$ under perturbations with magnitude $\delta$ in terms of a performance function $a(s,\delta)$ and a viability threshold $\tau$. 
Define
\begin{equation}
\label{eq:f-definition}
f_\tau(s, \delta) = \begin{cases}
a(s, \delta) & a(s,\delta) \geq \tau \\
0 & a(s,\delta) < \tau
\end{cases}
\end{equation} 
Substituting this definition of $f$ into \eqref{walshequation} yields
\begin{equation}
R_\tau(s) = \int a(s,\delta)\ \mathbb{I}(a(s,\delta) \geq \tau)\ p(\delta)\ d\delta
\end{equation}
Calculating this metric precisely requires specifying a probability distribution over perturbation sizes. 
In the absence of use-case-specific prior probability information, a sensible reference estimate can be obtained by using a uniform density over a sufficiently large perturbation range $[0,M]$ such that no classifier under consideration would be viable with perturbations larger than $M$.
\begin{align}
R_\tau(s) &= \int_0^M a(s,\delta)\ \mathbb{I}(a(s,\delta) \geq \tau)\ M^{-1} \ d\delta
\end{align}
Because the integrand is zero for $\delta$ above a threshold, the specific choice of $M$ does not affect the relative performance of classifiers so long as it is sufficiently large, as it acts only as a scaling constant. 
For the sake of standardization and simplicity, we drop the constant $M^{-1}$ and replace the integral bounds with $[0,D_{\tau}(s)]$, where $D{\tau}(s) = \inf \{\delta: a(s,\delta) < \tau\}$. 
This yields the EVP:
\begin{align}
 \label{eq:EVP}
 \mathrm{EVP}_a(s; \tau) &= \int_0^{D_{\tau}(s)} a(s,\delta)\ d\delta 
\end{align}

\begin{figure}[t]
    \centering
    \includegraphics[width=8cm]{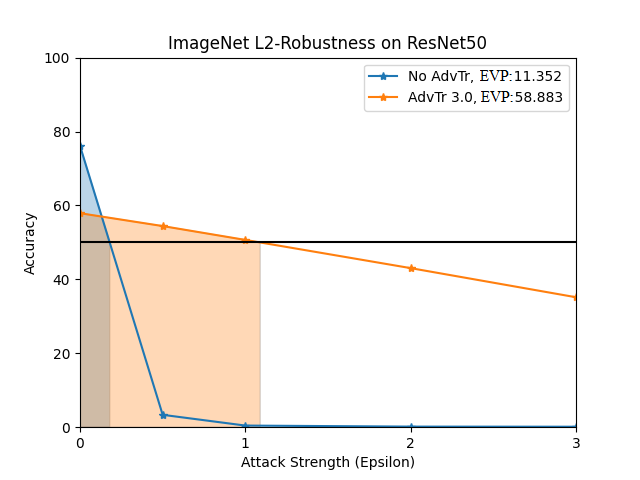}
    \caption{The Expected Viable Performance illustrated as area under an $L2$ Accuracy-Perturbation Curve on ImageNet. The viability threshold $\tau$ is set to 50\% accuracy.}
  \label{fig:manip}
\end{figure}

Equation~\eqref{eq:EVP} is simply the area under the Performance-Perturbation Curve over the perturbation range that yields viable performance;
  it is equivalent to the expected functionality formulation if perturbation magnitudes are scaled to the interval $[0,1]$. 
In Fig.~\ref{fig:manip}, we illustrate the EVP for two ResNet50 models, using accuracy as the performance measure and $\tau = 0.50$.
The blue area represents the EVP robustness quantity for the model without adversarial training, showing a steep decline in performance and lower EVP relative to the adversarially trained ResNet50 (orange) demonstrating graceful degradation and a larger EVP, consistent with a more robust classifer.

\begin{figure}[t]
    \centering
    \includegraphics[width=8cm]{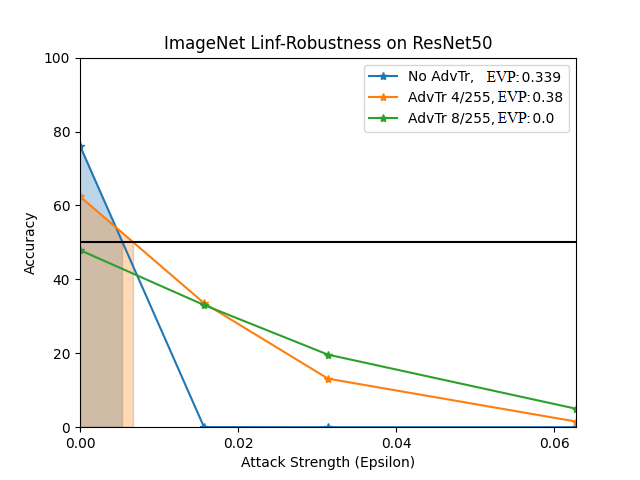}
    \caption{EVP illustrated as area under an $L_\infty$ Accuracy-Perturbation Curve on ImageNet. Each $L_\infty$ value is reported divided by 255. $\tau=50\%$}
  \label{fig:imagenetinf}
  \end{figure}
  
Although the choice of $\tau$ should be motivated by concrete cost-benefit concerns for a particular application, a default value should depend at least on the number of classes involved in the application. 
As a simple default, a criterion based on achieving a target statistical effect size relative to a chance baseline can be used. 
For example, using a target Cohen's $d$ value to derive an accuracy threshold yields an expression for $\tau$ given by \eqref{eq:threshold-by-cohens-d}:
\begin{equation}
\label{eq:threshold-by-cohens-d}
\tau = \frac{1}{C} + d\sqrt{\frac{1}{C}\left(1-\frac{1}{C}\right)}
\end{equation}
Using a target $d = 0.5$ yields the accuracy thresholds expressed as percentages in Table~\ref{tab:cohen}.

\begin{table}
    \caption{Suggested $\tau$ leveraging Cohen's $d$}
    \label{tab:cohen}
\begin{center}
  \begin{tabular}{|c c|} 
   \hline
   Number classes ($C$) & Performance Threshold ($\tau$) \\ [0.5ex] 
   \hline\hline
   2 & 75\% \\ 
   \hline
   5 & 40\% \\
   \hline
   10 & 25\% \\
   \hline
   100 & 6\% \\
   \hline
   1000 & 1.7\% \\ [1ex] 
   \hline
  \end{tabular}
  \end{center}
\end{table}

\subsection{Approximate Calculation}
Calculating EVP in practice requires evaluating the accuracy of a classifier under perturbations of varying magnitudes. 
Test items are generated subject to perturbations of a variety of sizes from $0$ to $M$, where $M$ represents the largest perturbation at which some classifier is considered viable. 
An increasing sequence of perturbation magnitudes, $\epsilon_1, \dots, \epsilon_N$, can be used, and the EVP integral can be approximated via a Riemann sum, for example, using the ``trapezoid method'':
\begin{align}
\textrm{EVP}_a(s; \tau) = \sum_{i=1}^{N} \frac{f(s, \epsilon_i) + f(s, \epsilon_{i-1})}{2} (\epsilon_i - \epsilon_{i-1})
\end{align}
where $\epsilon_0 = 0$, $f$ is defined as in \eqref{eq:f-definition}, and $f(s,\epsilon_0)$ is the ``clean'' performance of the classifier.

This definition of robustness takes into account model functionality given a perturbation as well as the size of the perturbation space itself.
A model that maintains performance at increasing levels of perturbation is similarly described in earlier robustness metrics such as the minimum perturbation distance and the ARA score~\cite{woods}, whereas metrics such as adversarial accuracy are specific to one level of perturbation.
We argue that EVP is a useful addition to the robustness toolbox because it takes into account viability as well as the magnitude of the perturbation space.

\subsection{Relationship to Accuracy-Robustness Area}
The robustness metric most similar to EVP is the ARA~\cite{woods} which also aggregates performance across perturbation sizes using the area under a performance curve. 
EVP differs in two ways from the ARA, however. 
The first is that the ARA takes into account perturbations large enough to produce chance classification performance, even if a classifier would have been rendered unusable under much smaller perturbations.
EVP only credits a classifier with robustness against perturbations that yield performance above a minimum \emph{practical} threshold.
Therefore, EVP lends itself to practical application and accreditation of models for target uses.

The two metrics become more similar if we set $\tau = 1/C$, or modify the ARA to compare to a better performing baseline. 
In this case, both metrics are considering the same range of perturbations. 
However, whereas the ARA is calculated using the area between the performance of the classifier in question and that of the na\"{i}ve classifier, our metric uses the full area under the performance curve within the relevant perturbation interval. 
The difference is equal to the area of the rectangle with height $1/C$ and width equal to the perturbation magnitude that yields performance equal to the na\"{i}ve baseline. 
This difference will therefore be larger for classifiers with larger perturbation tolerances.
The practical effect is that, even after matching the thresholds, the ARA places relatively more emphasis on achieving better performance with small perturbations and less emphasis on the range of the perturbations that can be tolerated.

\section{Experiments}
\label{sec:experiments}

We demonstrate EVP on a series of attacks on ResNet50 models.
The models were trained on CIFAR (10 classes) and ImageNet (1000 classes) datasets.
We measured perturbations with the $L_2$ norm. 
Pretrained ResNet50 models provided by the \emph{robustness} library~\cite{robustness} were used for all experiments. 

\subsection{Impact of Adversarial Training on EVP}

\begin{figure}[ht]
  \centering
  \includegraphics[width=8cm]{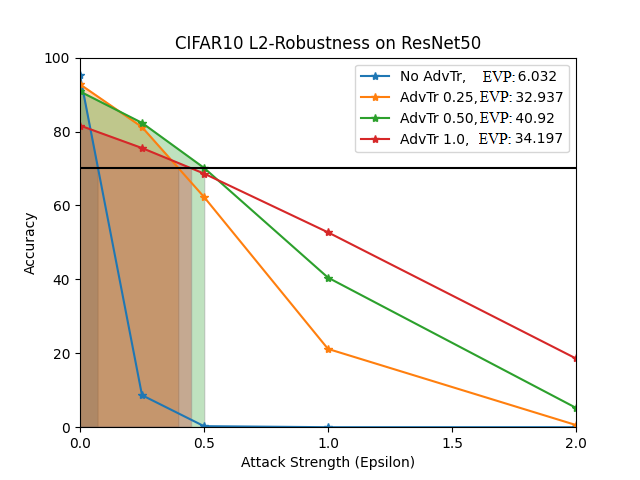}
  \caption{Adversarial training increases the robustness of the model up to a point. Because EVP discounts performance gains below a pre-defined viability threshold, training to defend against large perturbations produces a net decrease in robustness because the gains are accruing below the threshold, and are paid for by losses above it. Sampling interval is 0.25 $L_2$ distance.}
\label{fig:cifar}
\end{figure}

In Fig.~\ref{fig:cifar}, we show that classifiers fit using adversarial training from \cite{robustness} at a 0.5 $L_2$ bound (green) produced a more robust model according to EVP with $\tau = 0.70$ than larger $L_2$ distances (red). 
This happens because training the classifier to be robust to larger perturbations causes it to forfeit some accuracy against smaller perturbations. 
While a metric like ARA would consider this tradeoff a net win in this case, EVP does not, because the gains are accruing below the viability threshold and are therefore discounted.

\begin{figure}[t]
  \centering
  \includegraphics[width=8cm]{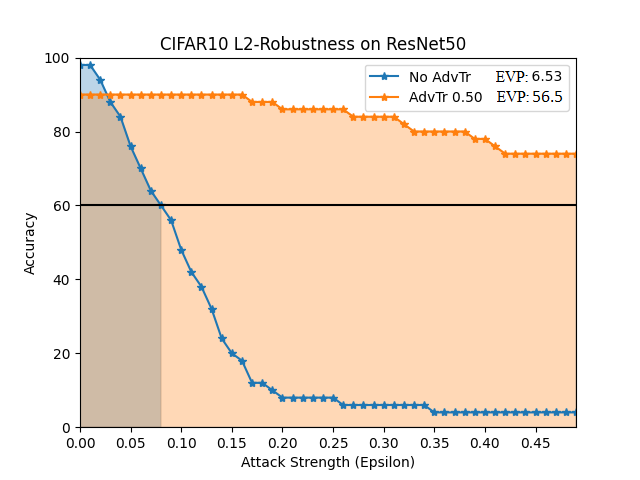}
  \caption{Using a smaller interval of 0.01, $L_2$ distance calculates an $\textrm{EVP}=6.5$ on the naturally trained model of CIFAR. Due to the visualization of the blue region, the EVP of the orange region is incompletely plotted.}
\label{fig:convergenceplot}
\end{figure}

This method improves on adversarial accuracy because it takes into account all of the potential adversarial attack bounds. 
However, it is significantly more computationally expensive due to the number of adversarial accuracies that must be computed. 
It is recommended that the interval size be small enough to distinguish among classifiers and to yield a smooth performance-perturbation curve when plotted.
Any obvious bumps in the curve may require a locally finer resolution in that region of the curve.

\subsection{Influence of $\tau$ on Choice of Loss Function}

The flexibility of the EVP to be defined for different viability thresholds allows a user to customize the metric to their application. 
While one model may outperform another at one viability threshold, the reverse may hold at a different threshold. 
In particular, a stringent (high) threshold for viability will incentivize high-performance on natural data, while a lower threshold will create a greater incentive for ``graceful degradation'' over a large perturbation range. 

\begin{figure}[t]
  \centering
  \includegraphics[width=8cm]{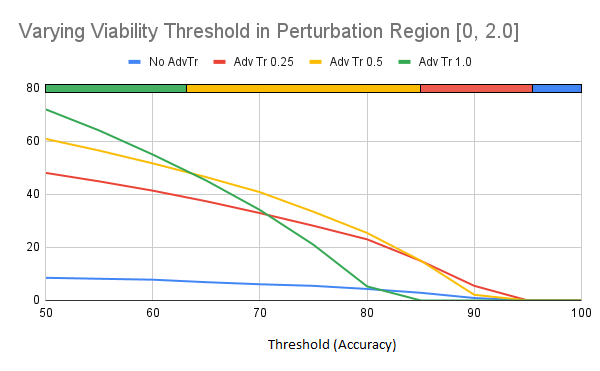}
  \caption{The EVP will vary with threshold, which is chosen based upon a given application. System viability affects the measurement of robustness. The most robust model given a threshold is indicated via the colored segments at the top.}
\label{fig:varythreshold}
\end{figure}

In Fig.~\ref{fig:varythreshold}, a natural CIFAR model is compared with three adversarially trained models. 
At a viability threshold above around 95\% accuracy, the naturally trained model outperforms the three adversarially robust models as measured by EVP. 
For viability thresholds between 85\% and 95\%, the model trained to be robust against small (0.25) adversarial perturbations pulls ahead. 
As the threshold continues to decrease, the optimal training regimen involves successively larger adversarial perturbations. 
This progression reflects the trade-off between performance on clean data and robustness to large perturbations.

\subsection{Convergence of Trapezoid Approximation}
\begin{figure}[t]
  \centering
  \includegraphics[width=8cm]{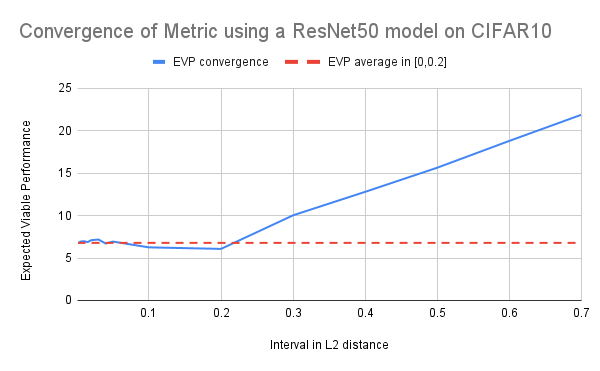}
  \caption{Selecting the interval size in the perturbation space is stable at $\Delta\epsilon < 0.2$ in $L_2$ distance.}
\label{fig:convergence}
\end{figure}

An important part of any metric is that it converges to a single value given greater levels of precision. 
We demonstrate via experiments that EVP converges with more sampling intervals in the perturbation space.
In Fig.~\ref{fig:convergence}, finer intervals of the perturbation space allow for more precision in measuring EVP. 
One caveat to the measurement is understanding that the interval can only be as fine grained as the attack itself. 
A PGD step size must be less than the sampling interval. 
Using smaller intervals requires more adversarially attacked images to be computed for each interval, increasing the data storage and training time requirements. 
It is recommended to use an interval size of less than or equal to 0.1 $L_2$ distance. 
The PGD step size for Fig.~\ref{fig:convergence} is $step=0.005$ for all trials.

\section{Discussion}
\label{sec:discussion}
We presented a new formulation of ML robustness quantification, EVP, applicable to both adversarial and natural perturbations. 
The evaluation is based a domain-general approach to robustness quantification, previously demonstrated in the cognitive sciences~\cite{walshq}.  
EVP is a consistent measurement with previous conceptions of robustness, but introduces a threshold to define functionality in a specific environment. 
The goal of this research is to provide a formulation of robustness that can be applied in evaluating models in an applied setting where functionality is more important than performance alone.

\subsection{Generalizations}
\paragraph{Choice of the performance metric}
Though we have focused on performance defined by 0-1 accuracy here, metrics such as precision, recall, and the F1-score can easily be substituted to define functionality, with the same considerations that would usually motivate the use of such metrics, including imbalanced training sets, asymmetric misclassification costs, etc.
EVP may be used to quantify robustness with multiple performance metrics of interest within a single application (multivariate functionality).
It allows for variation in performance metrics across settings (application, model, data types, decision type) while maintaining a consistent interpretation of robustness.

\paragraph{Natural perturbations}
EVP applies across any source of perturbation, as long as we can define the range of relevant perturbation values.
Robust models are desirable not only in the face of potential adversarial attacks, but also in noisy test environments or where data is subject to more natural ``corruption'', than that from which training data is typically derived. 
Here, we focused on adversarial perturbations and reported the robustness along with the kind of attack. 
However, the same robustness methodology can be used for any attack or corruption vector and any measure of performance.

\paragraph{Mechanisms of robustness}
By separating robustness quantification from the source of perturbations, EVP allows researchers to test potential mechanisms and methods that may contribute to ML robustness.
We refer to these as the \emph{mechanisms of robustness}, elements of the ML system or training processes that contribute to the functionality, stability, or robustness.
Adversarial training is a mechanism of robustness, as is active learning.
As illustrated in Figs.~\ref{fig:cifar} and~\ref{fig:convergenceplot}, for example, EVP allows direct assessment and comparison of mechanisms, enabling inferences about how degree and kind of mechanism may increase or decrease ML robustness.

\paragraph{Perturbation threshold}
A threshold in the perturbation space may have a similar effect such that large perceptual changes in the image should not be included in the evaluation. A model is functional when it evaluates non-perceptual perturbations as the same class as the original image. Perceptual perturbations may be semantically meaningful and thus labeling them as adversarial perturbations may contradict the intended functionality of a model.

\subsection{Future Work}
The metric is useful for applications where the user has an understanding of the requirements for the system's functionality. When it comes to implementation of systems and evaluating the robustness of those systems for real world scenarios, the metric quantifies robustness to small adversarial perturbations while discounting gains accruing below a useful level of functionality, which are also more likely to be illusory due to representing semantic changes. In future research we intend to pursue methods for determining the threshold of viability, as well as generalizations of EVP involving "soft" viability thresholds.


\section{Acknowledgements}
The opinions expressed herein are solely those of the authors and do not necessarily represent the opinions of the United States Government, the U.S. Department of Defense, the Department of the Air Force, or any of their subsidiaries or employees.
This research was funded by a Commander's Research and Development Fund grant from the Air Force Research Laboratory.
DISTRIBUTION A: Cleared for public release, distribution unlimited (AFRL-2023-0477).

\bibliographystyle{IEEEtran}
\bibliography{main}

\end{document}